\documentclass[sigconf]{acmart}
\usepackage{amsfonts}
\usepackage{amsmath}
\usepackage{graphicx}
\usepackage{tabulary}
\usepackage{multirow}
\usepackage{caption}
\usepackage{subcaption} 
\usepackage{makecell}
\AtBeginDocument{%
  \providecommand\BibTeX{{%
    \normalfont B\kern-0.5em{\scshape i\kern-0.25em b}\kern-0.8em\TeX}}}

\setcopyright{acmcopyright}
\copyrightyear{2018}
\acmYear{2018}
\acmDOI{XXXXXXX.XXXXXXX}

\acmConference[Conference acronym 'XX]{Make sure to enter the correct
  conference title from your rights confirmation emai}{June 03--05,
  2018}{Woodstock, NY}
\acmPrice{15.00}
\acmISBN{978-1-4503-XXXX-X/18/06}

\begin{document}

\title{GreenEyes: An Air Quality Evaluating Model based on WaveNet}

\author{Kan Huang}
\email{kan.huang@connect.ust.hk}
\affiliation{
  \institution{The Hong Kong University of Science and Technology}
  \streetaddress{Clearwater Bay}
  \city{Hong Kong}
  \country{China}
}

\author{Kai Zhang}
\email{kaz321@lehigh.edu}
\affiliation{
  \institution{Lehigh University}
  \streetaddress{27 Memorial Dr W, Bethlehem, PA 18015}
  \city{Bethlehem}
  \country{United States}
}

\author{Ming Liu}
\email{eelium@ust.hk}
\affiliation{
  \institution{The Hong Kong University of Science and Technology}
  \streetaddress{Clearwater Bay}
  \city{Hong Kong}
  \country{China}
}


\begin{abstract}
Accompanying rapid industrialization, humans are suffering from serious air pollution problems. The demand for air quality prediction is becoming more and more important to the government's policy-making and people's daily life. In this paper, We propose GreenEyes -- a deep neural network model, which consists of a WaveNet-based backbone block for learning representations of sequences and an LSTM with a Temporal Attention module for capturing the hidden interactions between features of multi-channel inputs. To evaluate the effectiveness of our proposed method, we carry out several experiments including an ablation study on our collected and preprocessed air quality data near HKUST. The experimental results show our model can effectively predict the air quality level of the next timestamp given any segment of the air quality data from the data set. We have also released our standalone dataset at \href{https://github.com/AI-Huang/IAQI_Dataset}{this URL}. The model and code for this paper are publicly available at \href{https://github.com/AI-Huang/AirEvaluation}{this URL}.

\end{abstract}



\maketitle

\section{Introduction}

With the development of the global economy and industrialization, people's living standards have improved, in the meanwhile, environmental problems such as air pollution have become a big concern. As World Health Organization (WHO) stated \cite{world2016ambient}, air pollution is the world's largest environmental health risk, which will incur many diseases including but not limited to respiratory infections, heart disease, COPD, stroke, and lung cancer. Among all kinds of pollution, air pollution has the largest impact on premature deaths annually \cite{lelieveld2015contribution}. Hence, as people's awareness of health increases, more and more smart devices such as smart bands have been developed and equipped, which can report air quality status. Moreover, a smart indoor air purifier can automatically purify the air when the resident is not at home.

\begin{figure}[!htbp]
    \begin{center}
    \includegraphics[width=\linewidth]{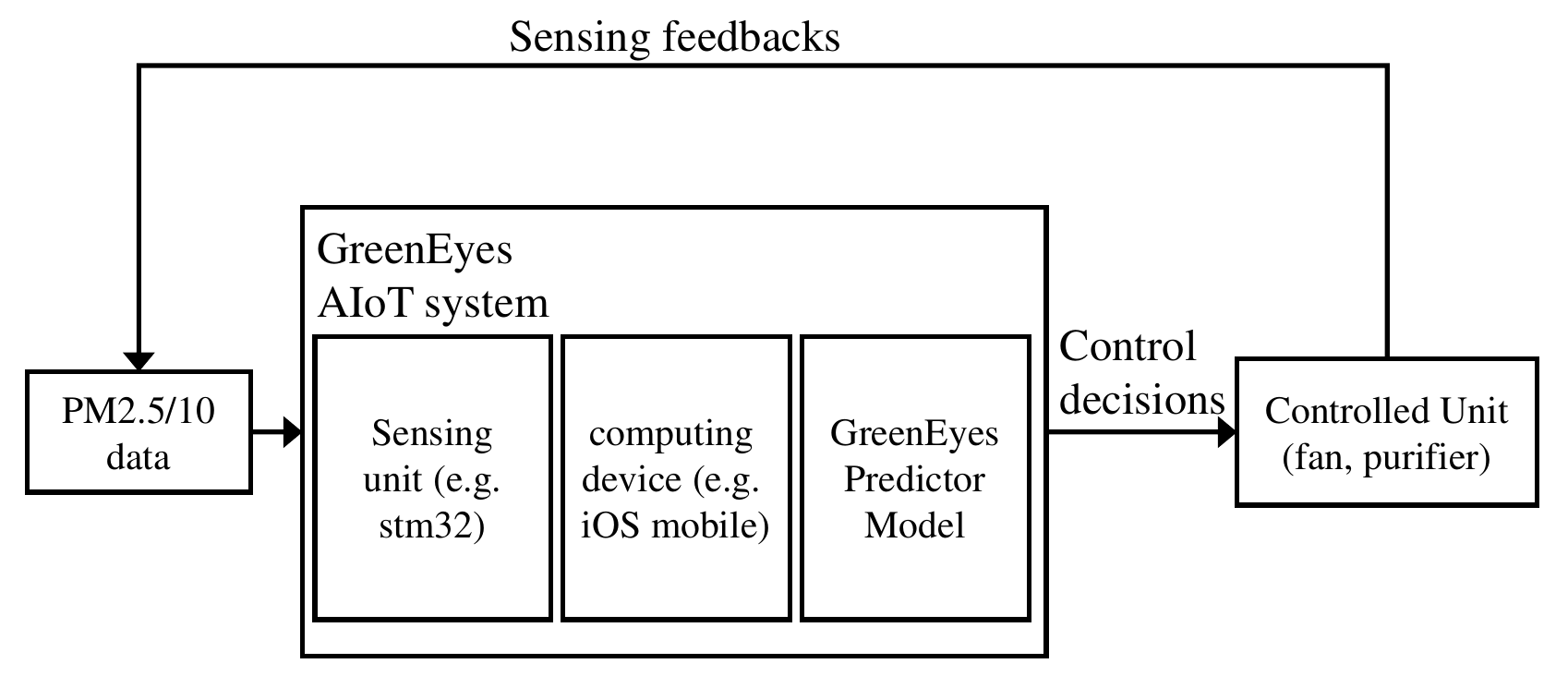}
    \end{center}
    \caption{GreenEyes: AIoT deployment.}
    \label{fig:greeneyes_aiot}
\end{figure}

The air pollution problem is widely discussed in the field of Artificial Intelligence of Things (AIoT) and Sensing Networks. Some IoT systems with variant functions are designed to monitor air quality for different application scenarios \cite{kumar2017air, oh2015indoor, zheng2016design}. For instance, Ray et al. \cite{ray2016internet} built a smart air-borne PM2.5 density monitoring system based on the cloud platform. However, these systems simply execute quality detection tasks without considering future air quality to let the purifier intelligently control its power level for energy-saving purposes. To bridge this gap, we propose the GreenEyes framework to predict the trend with previous air pollution levels. The feedback control system can be illustrated as Figure \ref{fig:greeneyes_aiot} shows.

In this work, we firstly investigate the problem of preprocessing noisy PM2.5 sequence data and creating an appropriate supervising target sequence. We implement the GreenEyes model to predict the future air quality and evaluate it on each channel of PM2.5 data. Besides, we train our model with all channels' data together. Other works either use different kinds of data \cite{han2020joint}, or use sensors of the same model but place them at different places \cite{ray2016internet}. The former methodology is Multi-sensor Fusion \cite{wang2019multi}, it is widely used in the intelligent and autonomous systems \cite{luo1989multisensor, hall1997introduction, wang2012towards, cai2020probabilistic}. However, our approach of experiment proves that multi-sensors (of the same model, at the same place) will make the model perform better in predicting target data.

The main characteristics of this paper are summarized as follows:
\begin{itemize}
\item We treat WaveNet's residual layers as a feature block. This idea comes from the basic structures such as convolution-activation-pooling in computer vision. Such a design can increase reception filed and learn better representations.
\item We innovatively stack several WaveNet blocks to build the model's main body. As the basic mechanism of deep learning networks is to build models brick by brick, the same module with different parameters is usually used in the same model. We borrow this idea and make it possible to parameterize our model. The model's optimal hyperparameters such as depth and filters can also be fine-tuned easily.
\item We put Attention \cite{bahdanau2014neural} and LSTM \cite{hochreiter1997long} at the endpoint as output layers. Ablation experiments demonstrate its necessity because this module can capture the hidden interations between features of different sequences (channels).
\end{itemize}


\section{Datasets}\label{sec:data_modeling}

AQI (\textbf{A}ir \textbf{Q}uality \textbf{I}ndex) is widely used for measuring the current pollution status of the air. IAQIs (\textbf{I}ndividual \textbf{A}ir \textbf{Q}uality \textbf{I}ndex) are calculated according to pollutants such as ozone, nitrogen dioxide, sulphur dioxide, and others, before final AQI is concluded. In our work, the IAQI of $PM_{2.5}$ is considered.

IAQI level data calculated from raw air quality data of sensors cannot be used directly because of high-frequency noise. As Figure \ref{fig:pm25_0_origin_and_labeled_level} presents, in some intervals of the time axis, the IAQI level fluctuates very fast. This is because the air quality data is exactly fluctuating around the threshold line. In real AIoT applications, we don't need this fluctuation. Image the following module is a fan switch that takes the model's output to determine and we want this output to be relatively stable. In order to clean the data fluctuation while keeping the trend features, we innovatively brought out a method of human manually labeling. It creates an appropriated target label function that the model can learn. Also, based on the labeling tricks, the problem that the predictions on the IAQI level will fluctuate near the thresholds is much reduced.

\subsection{Data Collection}

We placed our 4 sensors in an office room located inside the Academic Building of the HKUST. The room is inside the academic building and has no windows, it provides a stable experimental environment for temperature and humidity. The sampling rate of the sensor is 1 Hz. We simultaneously collected around 220k data points for each sensor in a continuous period starting from 20:28 on 25th November 2019. This period is about 2 days and a half or 61 hours.

\subsection{IAQI Calculation}

The final AQI depends on each pollutant's IAQI, which is calculated by Equation \ref{eq:IAQI_p}

\begin{equation}
    \label{eq:IAQI_p}
    IAQI_p = \frac{C_p-BP_{Lo}}{BP_{Hi}-BP_{Lo}}(IAQI_{Hi}-IAQI_{Lo})+IAQI_{Lo},
\end{equation}

and finally, AQI is calculated by Equation \ref{eq:AQI}
\begin{equation}
    \label{eq:AQI}
    AQI = \text{max}\{IAQI_{1}, IAQI_{2}, IAQI_{3}, ..., IAQI_{n}\}.
\end{equation}

In this paper, we only concern and discuss on the IAQI regarding $PM_{2.5}$.



Above equations about IAQI and AQI are universal for multi kinds of air pollution standards. Different thresholds are used when mapping air pollutants data into IAQI in different standards. Table \ref{table:IAQI_thresholds} lists $PM_{2.5}$ and $PM_{10}$ IAQI thresholds in China's and USA's standards respectively. In this paper, we use the \textbf{USA standard}. 

\begin{table}[!htbp]
    \centering
    \caption{Concentration thresholds of IAQI w.r.t. pollutant categories, USA}
    \label{table:IAQI_thresholds}
    \begin{tabular}{l|l|l|l|l}
        \hline
        \hline
        \  & USA & USA & China & China \\ \hline
        IAQI & \makecell[c]{$PM_{2.5}$ \\ ($\mu g/m^3$)} & \makecell[c]{$PM_{10}$ \\ ($\mu g/m^3$)} & \makecell[c]{$PM_{2.5}$ \\ ($\mu g/m^3$)} & \makecell[c]{$PM_{10}$ \\ ($\mu g/m^3$)} \\ \hline
        0    & 0     & 0   & 0   & 0   \\ \hline
        50   & 12.1  & 55  & 35  & 50  \\ \hline
        100  & 35.5  & 155 & 75  & 150 \\ \hline
        150  & 55.5  & 255 & 115 & 250 \\ \hline
        200  & 150.5 & 355 & 150 & 350 \\ \hline
        300  & 250.5 & 425 & 250 & 420 \\ \hline
        400  & N/A   & N/A & 350 & 500 \\ \hline
        500  & 500.4 & 604 & 500 & 600 \\ \hline
        \hline
    \end{tabular}
\end{table}

\begin{figure*}[!htbp]
    \begin{center}
        \includegraphics[width=0.8\linewidth]{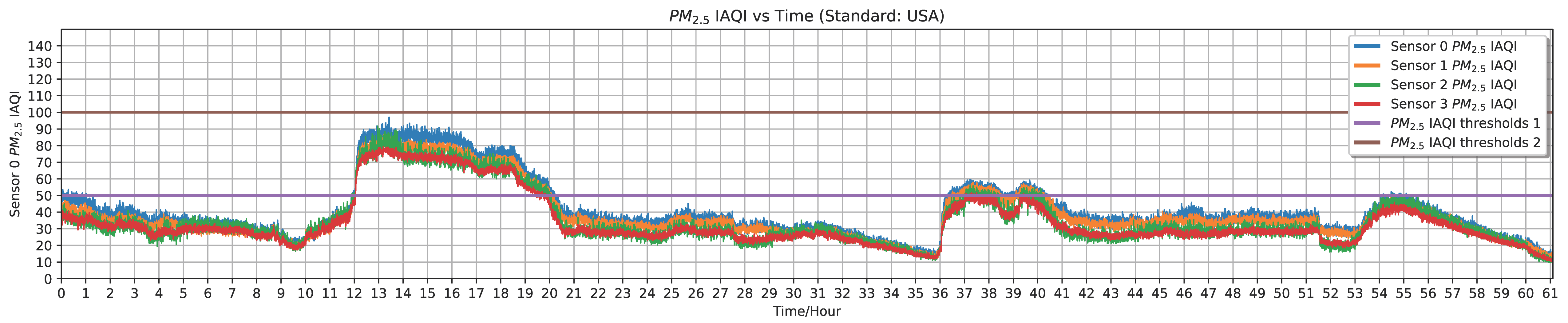}
    \end{center}
    \caption{All $PM_{2.5}$ data with IAQI thresholds.}
    \label{fig:pm25_all_iaqi_with_thresholds}
\end{figure*}

Figure \ref{fig:pm25_all_iaqi_with_thresholds} shows the IAQI curves corresponding to $PM_{2.5}$, with thresholds lines.

\subsection{Data Polynomialization}

The task of our model is to predict the IAQI level when inputting a segment of air pollutant concentration data. However, the origin IAQI level lines cannot be directly used because 1. in deep learning, a step function is very hard to learn especially on the rising and falling edges; 2. in some areas the IAQI level fluctuates extremely frequently, which makes the learning even harder, as shown in Figure \ref{fig:pm25_0_origin_and_labeled_level}.


We're inspired by B. Rouet-Leduc's work \cite{rouet2017machine} on earthquake predicting, where the earthquake events are represented as failures. The target function is designed as a series of descending ramps. Mathematically, these descending ramps form a polygonal function and count down the time to next earthquake. The problem is turned into fitting the time counting down curve with the acoustic data input.

A polygonal function, also named piece-wise linear function $f(x)$, is a continuous function mathematically defined on an interval $[a, b]\in \mathbb R$ such that $[a, b]$ can be divided into a set of intervals on each of which the function is a linear function, that is, there exists a subdivision 
\begin{equation}
    a=x_0 < x_1 < ... < x_n=b
\end{equation}
such that $f(x)$ is linear on each interval $[x_{n-1}, x_n]$.

Polygonal functions can be used to generate approximations to known curves, planes, etc. Also, for unknown data, polygonal functions can also be learned by some algorithms such as decision tree, to fit the data. In our work of predicting, polygonal functions help us to eliminate the \textbf{hesitation} area, and build the target data.

\subsection{Data Polygonalization: Human Labeling based on Decisions}

\begin{figure*}[!htbp]
    \begin{center}
        \includegraphics[width=0.8\linewidth]{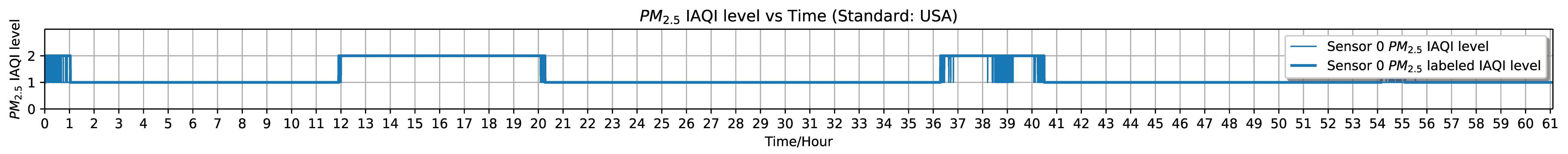}
        \caption{Sensor 0's $PM_{2.5}$ origin and labeled IAQI level.}
        \label{fig:pm25_0_origin_and_labeled_level}
    \end{center}
\end{figure*}

We firstly label by hand the level step down\/up points, and map them into rising\/falling lines. This method transfer discrete decision points into continuous target data series which have the same dimension as the time indices and corresponding $PM_{2.5}$ data. This kind of method make us get the polygonal target data as B. Rouet-Leduc, et al. \cite{rouet2017machine} did. Figure \ref{fig:pm25_0_origin_and_labeled_level} shows our labeling results.

\begin{figure*}[!htbp]
    \centering
    \includegraphics[width=0.8\linewidth]{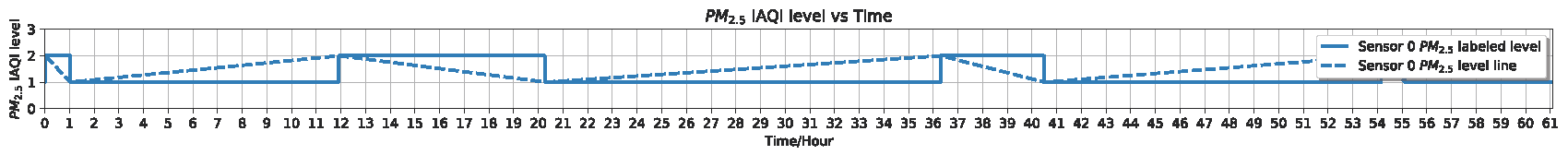}
    \caption{Sensor 0's labeled $PM_{2.5}$ IAQI level and its polygonal line.}
    \label{fig:pm25_0_labeled_level}
\end{figure*}

Finally, the labeled target data is turned into a polygonal function. Dash lines in Figure \ref{fig:pm25_0_labeled_level} shows the results. These triangular lines will be label data for our supervised learning fitting problem.

The level polygonal lines w.r.t. their manually labeled level curves can be written as equations with form below
\begin{equation}
    \label{formula:polygonal}
    L_i(t)=k_i*t+b_i,\ where\ t\in[t_i,t_{i+1}],
\end{equation}
where $k_i$ is the slope of the curve, $t_i$ and $t_{i+1}$ are start and end time point for every interval of the polygonal line. When $k_i>0$, the trend of the IAQI level is raising, and vice versa. The absolute value of $k_i$ is the approximate and potential changing speed of IAQI level. Thus, every polygonal line can be divided into several segments within time interval $t_i$ to $t_{i+1}$, and every segment estimates the 1-order approximate trend w.r.t. original IAQI level within corresponding time interval. For the i-th segment, $k_i$ is
\begin{equation}
    k_i = \frac{l_{i+1}-l_i}{t_{i+1}-t_i}.
\end{equation}
where $l_{i+1}$ and $l_i$ are the original IAQI levels at end and start time $t_{i+1}$ and $t_i$.

Our experiments will take these polygonalized IAQI level lines as the supervising data. The fitting problem can be described as: given a IAQI sequence of \textit{windows size}, predict the IAQI level of the next time frame after this time window.

\section{Methodology}\label{sec:model_design}

\begin{figure*}[!htbp]
  \centering
  \includegraphics[width=0.8\linewidth]{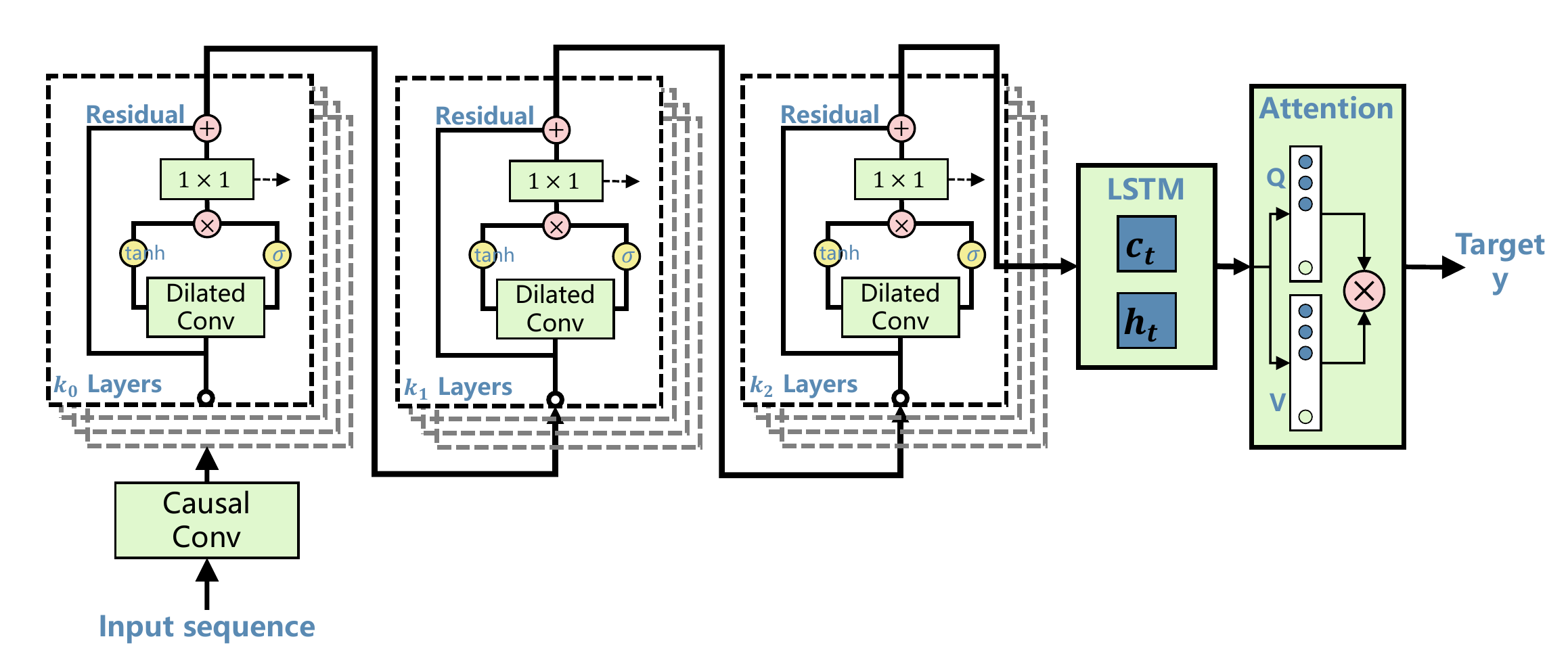}
  \caption{GreenEyes sequence to point fitting model.}
  \label{fig:greeneyes_model}
\end{figure*}

Recently, a series of neural networks related to the auto-regression model has been proposed and applied in regarding problems. DeepMind's WaveNet \cite{oord2016wavenet} is one of the famous and foundation work in between those \cite{shen2018natural}, \cite{wang2017tacotron} tackle with sequence representation and generating. For WaveNet's application scenario, the joint probability of the target sequence $\textbf{x}=\{x_1,x_2,...,x_T\}$ is factorized as a product of conditional probabilities as the below Equation \ref{eq:auto-regression-model}. Given an input $\bar{\textbf{x}}=\{x_1,x_2,...,x_{T-1}\}$ and with this conditional probabilities, we can obtain the distribution of the value $x_T$, and make generation samples.

\begin{equation}
    p(\textbf{x})=\prod_{t=1}^Tp\{x_t|x_1,x_2,...,x_{t-1}\}
    \label{eq:auto-regression-model}
\end{equation}

Auto-regression models can not only be used in data generation, but also in time series prediction. In our work, every sample $x_t$ and $y_t$ at any time step $t$ is conditioned on the samples at all previous timestamps, making it a multivariate auto-regression task. To limit the input length,  we only consider the conditional probabilities between $x_t$ and a sequence $x_{t-1-window\_size:t-1}$ with length $window\_size$. Different with other multivariate auto-regression tasks where sequences on all the temporal axis are modeled, we haven't used the sequence $y_{t-1-window\_size:t-1}$ to predict $y$, instead, we predict $y$ only with $x_{t-1-window\_size:t-1}$.

Different from B. Rouet-Leduc's work\cite{rouet2017machine}, in which random forest is used to predict seismic precursors, we use WaveNet as our GreenEyes model's main part. Air pollution data has the same structure as audio data. It is pretty suitable to utilize WaveNet as air pollution data can be modeled in the same way. Also, WaveNet's dilated causal convolutions and residual and skip connections are suitable for air pollution data.

We used the original WaveNet's core part as a WaveNet Block as we believe this block\-style configuration is more modularized, for we could change these blocks' hyper parameters more easily. Each WaveNet Block, as the same with WaveNet, contains several dilated convolution layers, called WaveNet Layer. Different dilation rates are also set on them, following DeepMind's original work. 

The designing of neural networks for deep learning has always followed principles such as modularization, and expandability. Well-known networks, such as VGG \cite{simonyan2014very} and ResNet \cite{he2016deep}, all have these features. VGG has two model types VGG16, and VGG19, with different model depth. And ResNet has models ResNet-18, ResNet-34, ResNet-50, etc. The cutting-edge model, Transformer \cite{vaswani2017attention}, also obeys these designs which makes it possible to build multi variant models for various sizes and application scenarios. Our model is designed for parameterization, too. Following our constructions, finally we set 8 WaveNet layers for the first block; and 5 layers for the second, 3 layers for the third. All blocks share the same kernel size of 3, and filters of 16. This set of hyper parameters are chosen by empiric and the computational capability of a 1080 Ti GPU. There might be more optimal parameters to search in future works.

As for the Attention layer, we set up two kinds of Attention mechanism - Dot-product attention layer, a.k.a. Luong-style attention \cite{luong2015effective}, as Equation \ref{eq:dot-product-attention} shows. We use the input for all value vector, key vector, and query input. Another mechanism is made by ourselves, called Temporal Attention. 

\begin{gather}
    \text{scores}=QK^T \\
    \text{Attention}(Q, K, V ) = \text{softmax}(QK^T)V
    \label{eq:dot-product-attention}
\end{gather}

In our Attention layer, we still use the Luong's multiplicative style attention \ref{eq:Luong-multiplicative-style} to gain score, but we simply it with a FC network. Moreover, we don't use softmax function to compute the attention weight. Rather, we use the function as Equation \ref{eq:temporal-attention} shows.

\begin{equation}
    \text{scores}(\bar {\textbf{\textit h}}_t^T,\bar {\textbf{\textit h}}_s)=\bar {\textbf{\textit h}}_t^T\textbf{\textit W}\bar {\textbf{\textit h}}_s 
    \label{eq:Luong-multiplicative-style}
\end{equation}

\begin{gather}
    \text{scores}=\textbf{\textit W}\textbf{\textit V}+\textbf{\textit b} \\
    \text{Attention}(V) = \text{exp}(\text{tanh}(\text{scores}))V
    \label{eq:temporal-attention}
\end{gather}

The reason that we replace the softmax with a tanh function followed by an exponential function, is to better adapt our model to the temporal data set. Our data set have many temporal and periodic features to learn. Tanh function is very common in sequential models, and it is also a component in every WaveNet layer.

\section{Experiments}

\subsection{Experimental Settings}

As we sampled $PM_{2.5}$ measurements from 4 sensors, Sensor 0 to Sensor 3, so we have a 4-channels $PM_{2.5}$ IAQI data set. Each channel's data can be taken as an individual data set. The stride is set as $\{10, 5, 2\}$, respectively. Besides, we fuse data from all channels to create a new data set named $PM2.5_{All}$.

Adam\cite{kingma2017adam} optimizer with an initial learning rate 0.0001 is applied in the experiments, which is multiplied by 0.1 after 20 epochs, where the total training epoch is 100.
We use mean squared error (MSE) and mean absolute error (MAE) as the evaluation metrics.


\subsection{Training and Validation} 

\subsubsection{Why did We Redesign the Attention Layer?}

At first, we utilized the dot-product attention layer provided by TensorFlow official. Table \ref{table:best_metrics_official_training} lists all the experiments' final best metrics during training.

\begin{table}[!htbp]
    \centering
    \begin{tabular}{c|c|c|c}
        \hline\hline
        Data & Stride & \makecell[c]{Minimum \\ train MSE} & \makecell[c]{Minimum \\ validation MSE} \\\hline
        \multirow{3}{*}{\text{$PM_{2.5}$(0)}} & 10 & 0.0969 & 0.1221 \\ \cline{2-4} 
                                        & 5 & 0.0071 & 0.0221 \\ \cline{2-4} 
                                        & 2 & 0.0049 & 0.0148 \\ \hline
        \multirow{3}{*}{\text{$PM_{2.5}$(1)}} & 10 & 0.0148 & 0.0226 \\ \cline{2-4} 
                                        & 5 & 0.0062 & 0.0137 \\ \cline{2-4} 
                                        & 2 & 0.0006 & 0.0039 \\ \hline
        \multirow{3}{*}{\text{$PM_{2.5}$(2)}} & 10 & 0.0087 & 0.0137 \\ \cline{2-4} 
                                        & 5 & 0.0080 & 0.0153 \\ \cline{2-4} 
                                        & 2 & 0.0027 & 0.0154 \\ \hline
        \multirow{3}{*}{\text{$PM_{2.5}$(3)}} & 10 & 0.0123 & 0.0209 \\ \cline{2-4} 
                                        & 5 & 0.0058 & 0.0110 \\ \cline{2-4} 
                                        & 2 & 0.0023 & 0.0037 \\ \hline
        \multirow{3}{*}{\text{$PM_{2.5}$(All)}} & 10 & 0.0182 & 0.0266  \\ \cline{2-4} 
                                        & 5 & 0.0039 & 0.0053 \\ \cline{2-4} 
                                        & 2 & 0.0010 & 0.0005 \\
        \hline
        \hline
    \end{tabular}
    \caption{Best metrics during training when applying official Attention.}
    \label{table:best_metrics_official_training}
\end{table}

\begin{figure}[!htbp]
    \centering
    \includegraphics[width=0.8\linewidth]{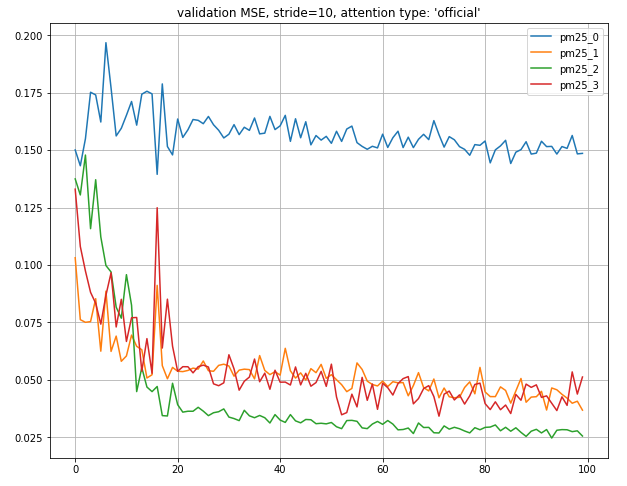}
    \caption{Validation MSE curves when applying official Attention ($stride=10$).}
    \label{fig:val_mse_official}
\end{figure}

\begin{figure}[!htbp]
    \centering
    \includegraphics[width=0.8\linewidth]{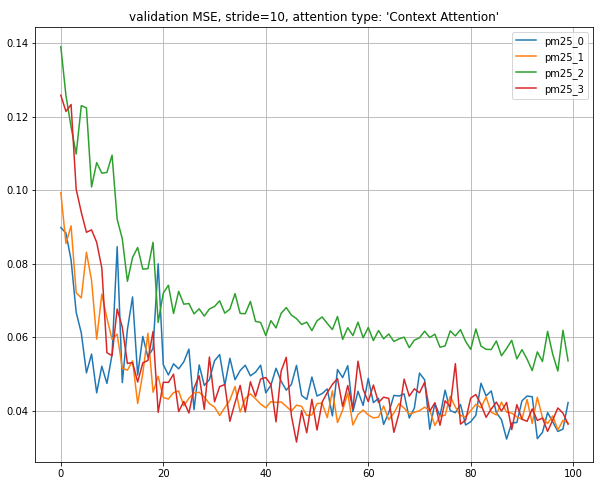}
    \caption{Validation MSE curves when applying Temporal Attention ($stride=10$).}
    \label{fig:val_mse_myattention}
\end{figure}

After we train the model with Temporal Attention, we discovery that the results on official Attention show limitations and defects. As Table \ref{table:best_metrics_myattention_training} shows, in most experiments, Temporal Attention outperforms official Attention. When we plot the validation curve, some principles can be figured out, specifically, Figure \ref{fig:val_mse_official} illustrates the validation MSE's curves with $stride=10$, and Figure \ref{fig:val_mse_myattention} illustrates the validation MSE's curves when applying Temporal Attention. We can conclude that when applying official Attention, the model cannot converge consistently with different data sets. Figure \ref{fig:val_mse_official} shows that model fails to converge when it learns on \text{$PM_{2.5}$(0)}. Meanwhile, applying Temporal Attention, the model can obtain a better MSE.

\subsubsection{Best Metrics during Training}



Table \ref{table:best_metrics_myattention_training} shows the experimental best metrics during training with the proposed Temporal Attention. It is obvious that in most cases, our Attention outperforms the official Attention. We also define a coefficient $ratio=\frac{min(train\ MSE)}{min(validation\ MSE)}$ to simply measure the generalization capability of the model.

\begin{table}[!htbp]
    \centering
    \begin{tabular}{c|c|c|c|c}
        \hline\hline
        Data & Stride & \makecell[c]{Minimum \\ train MSE} & \makecell[c]{Minimum \\ validation MSE} & ratio \\\hline
        \multirow{3}{*}{\text{$PM_{2.5}$(0)}} & 10 & 0.0223 & 0.0234 & 0.96 \\ \cline{2-5} 
                                        & 5 & 0.0034 & 0.0114 & 0.30 \\ \cline{2-5} 
                                        & 2 & \textbf{\textit{0.0006}} & \textbf{\textit{0.0035}} & 0.16 \\ \hline
        \multirow{3}{*}{\text{$PM_{2.5}$(1)}} & 10 & 0.0486 & 0.0510 & 0.95 \\ \cline{2-5} 
                                        & 5 & 0.0058 & 0.0142 & 0.41 \\ \cline{2-5} 
                                        & 2 & \textbf{\textit{0.0006}} & \textbf{\textit{0.0036}} & 0.17 \\ \hline
        \multirow{3}{*}{\text{$PM_{2.5}$(2)}} & 10 & 0.0171 & 0.0187 & 0.92 \\ \cline{2-5} 
                                        & 5 & 0.0024 & 0.0092 & 0.27 \\ \cline{2-5} 
                                        & 2 & \textbf{\textit{0.0012}} & \textbf{\textit{0.0066}} & 0.19 \\ \hline
        \multirow{3}{*}{\text{$PM_{2.5}$(3)}} & 10 & 0.0509 & 0.0468 & 1.09 \\ \cline{2-5} 
                                        & 5 & 0.0074 & 0.0167 & 0.44 \\ \cline{2-5} 
                                        & 2 & \textbf{\textit{0.0010}} & \textbf{\textit{0.0068}} & 0.15 \\ \hline
        \multirow{3}{*}{\text{$PM_{2.5}$(All)}} & 10 & 0.0068 & 0.0103 & 0.66 \\ \cline{2-5} 
                                        & 5 & 0.0014 & 0.0022 & 0.67 \\ \cline{2-5} 
                                        & 2 & \textbf{0.0007} & \textbf{0.0009} & 0.77 \\
        \hline
        \hline
    \end{tabular}
    \caption{Best metrics during training when applying Temporal Attention.}
    \label{table:best_metrics_myattention_training}
\end{table}

\subsection{Model Evaluation}

\begin{figure*}[!htbp]
    \centering
    \includegraphics[width=\linewidth]{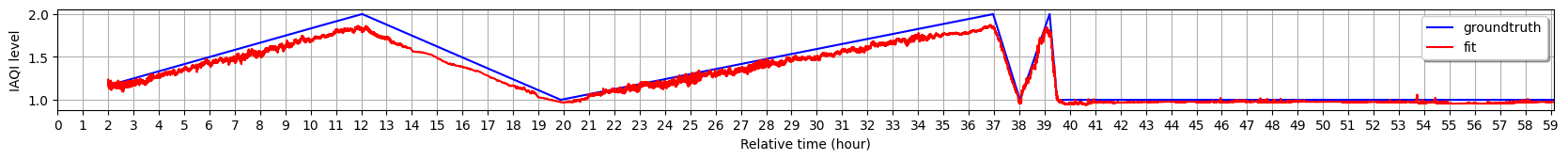}
    \caption{Evaluation of the GreenEyes model (\text{$PM_{2.5} (3)$}, stride=5).}
    \label{fig:model_eval_pm25_3_stride_5}
\end{figure*}

\begin{figure*}[!htbp]
    \centering
    \includegraphics[width=\linewidth]{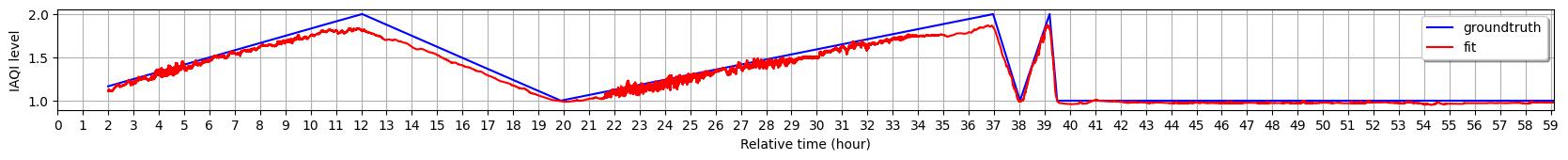}
    \caption{Evaluation of the GreenEyes model (\text{$PM_{2.5} (3)$}, stride=10).}
    \label{fig:model_eval_pm25_3_stride_2}
\end{figure*}

Figure \ref{fig:model_eval_pm25_3_stride_5} shows that our model fits the labeled IAQI level lines well, except that its predictions differ from the ground truth a little on some parts of the lines, especially on the turning corners. Figure \ref{fig:model_eval_pm25_3_stride_2} illustrates the same evaluation performance, which presents that the model may not need much data to learn as to set stride to 2.
To quantify the testing results of our model with different parameters, we test it on the whole $PM_{2.5}$ sequence by setting stride as 1. Table \ref{table:test_mse_mae} lists the statistics of our tests.


\begin{table}[!htbp]
    \centering
    \begin{tabular}{c|c|c|c}
        \hline\hline
        Data & Stride & MSE & MAE \\\hline
        \multirow{3}{*}{\text{$PM_{2.5}$(0)}} & 10 & 0.0266 & 0.13 \\ \cline{2-4} 
                                        & 5 & 0.0144 & 0.11 \\ \cline{2-4} 
                                        & 2 & \textbf{\textit{0.0037}} & \textbf{\textit{0.05}} \\ \hline
        \multirow{3}{*}{\text{$PM_{2.5}$(1)}} & 10 & 0.0517 & 0.18 \\ \cline{2-4} 
                                        & 5 & 0.0113 & 0.10 \\ \cline{2-4} 
                                        & 2 & \textbf{\textit{0.0036}} & \textbf{\textit{0.05}} \\ \hline
        \multirow{3}{*}{\text{$PM_{2.5}$(2)}} & 10 & 0.0188 & 0.11 \\ \cline{2-4} 
                                        & 5 & 0.0092 & 0.09 \\ \cline{2-4} 
                                        & 2 & \textbf{\textit{0.0069}} & \textbf{\textit{0.07}} \\ \hline
        \multirow{3}{*}{\text{$PM_{2.5}$(3)}} & 10 & 0.0501 & 0.16 \\ \cline{2-4} 
                                        & 5 & 0.0108 & 0.09 \\ \cline{2-4} 
                                        & 2 & \textbf{\textit{0.0070}} & \textbf{\textit{0.07}} \\ \hline
        \multirow{3}{*}{\text{$PM_{2.5}$(All)}} & 10 & 0.0118 & 0.09 \\ \cline{2-4} 
                                        & 5 & 0.0026 & 0.04 \\ \cline{2-4} 
                                        & 2 & \textbf{0.0010} & \textbf{0.02} \\ \hline
        \hline
    \end{tabular}
    \caption{Test MSE and MAE under different stride parameter.}
    \label{table:test_mse_mae}
\end{table}

\subsection{Ablation Study}

In order to validate the effectiveness of the modules, we conduct an ablation study on our GreenEyes model. We remove the bidirectional LSTM module and the mutli-head attention module, respectively, and get two model variance, w/o Attention and w/o LSTM. We plot the model's (w/o LSTM) training and validation curves as Figure and Figure show respectively.

\begin{figure}[!htbp]
    \centering
    \includegraphics[width=0.6\linewidth]{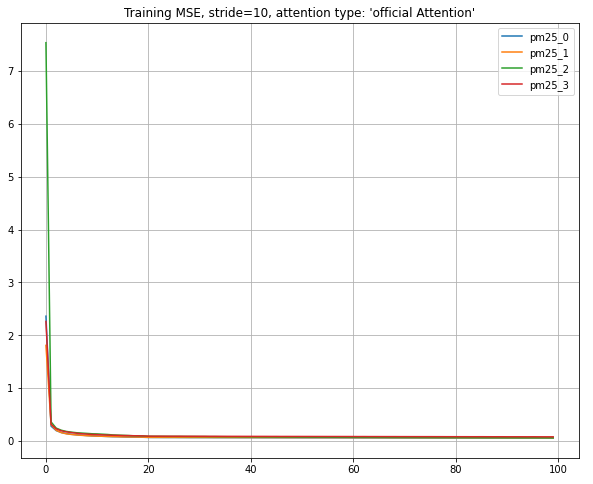}
    \caption{Model's (w/o LSTM) training plots (stride=10).}
    \label{fig:wo-LSTM-training}
\end{figure}

\begin{figure}[!htbp]
    \centering
    \includegraphics[width=0.6\linewidth]{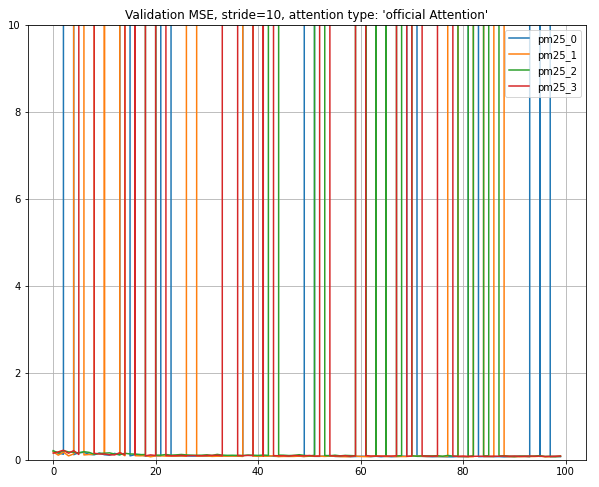}
    \caption{Model's (w/o LSTM) validating plots (stride=10).}
    \label{fig:wo-LSTM-validation}
\end{figure}

It is easily concluded that, without the LSTM layer, the model runs into the overfit status. Although it still fits well on the train set, it is rambling on the validation set.

In order to validate the Attention layer's function, we re-run the GreenEyes model with Temporal Attention on \text{$PM_{2.5}$(0)} to \text{$PM_{2.5}$(3)}, and then cut off this Attention layer and run the model again on the same data sets. Table \ref{table:myattention_vs_no_attention} shows the test MSE and MAE results of both configuration. It turns out that the model w/o Attention can perform better or is equivalent to the model applied with the Attention layer. However, by plotting the training curves again, we found that the model with the Temporal Attention layer can obtain smaller loss during training. 

\begin{table}[!htbp]
    \centering
    \begin{tabular}{c|c|c|c|c|c}
    \hline\hline
     & & \multicolumn{2}{c}{Our Attention} & \multicolumn{2}{c}{w/o Attention} \\ \cline{3-6}
    \multirow{-2}{*}{Data} & \multirow{-2}{*}{Stride} & MSE & MAE & MSE & MAE \\ \hline
    \multirow{3}{*}{\text{$PM_{2.5}$(0)}} & 10 & 0.0267 & 0.1438 & \textbf{0.0173} & \textbf{0.1189} \\ \cline{2-6}
      & 5 & 0.0119 & 0.1011 & \textbf{0.0092} & \textbf{0.0795} \\ \cline{2-6}
      & 2 & \textbf{0.0055} & \textbf{0.0661} & 0.0078 & 0.0721 \\ \cline{2-6}
    \multirow{3}{*}{\text{$PM_{2.5}$(1)}} & 10 & \textbf{0.0256} & 0.1383 & 0.0262 & \textbf{0.1292} \\ \cline{2-6}
     & 5 & 0.0151 & 0.1093 & \textbf{0.0097} & \textbf{0.0769} \\ \cline{2-6}
     & 2 & 0.0026 & 0.0423 & \textbf{0.0015} & \textbf{0.0305} \\ \cline{2-6}
    \multirow{3}{*}{\text{$PM_{2.5}$(2)}} & 10 & 0.0409 & 0.1548 & \textbf{0.0174} & \textbf{0.1170} \\ \cline{2-6}
     & 5 & 0.0116 & 0.0978 & \textbf{0.0053} & \textbf{0.0624} \\ \cline{2-6}
     & 2 & 0.0057 & 0.0577 & \textbf{0.0007} & \textbf{0.0202} \\ \cline{2-6}
    \multirow{3}{*}{\text{$PM_{2.5}$(3)}} & 10 & \textbf{0.0213} & \textbf{0.1338} & 0.0446 & 0.1726 \\ \cline{2-6}
     & 5 & \textbf{0.0064} & \textbf{0.0659} & 0.0083 & 0.0759 \\ \cline{2-6} 
     & 2 & 0.0044 & 0.0534 & \textbf{0.0018} & \textbf{0.0305} \\
    \hline\hline
    \end{tabular}
    \caption{Test MSE and MAE for model with and w/o Attention.}
    \label{table:myattention_vs_no_attention}
\end{table}

\subsection{Hyper-parameter Discussion}

\begin{figure}[!htbp]
    \centering
    \includegraphics[width=0.8\linewidth]{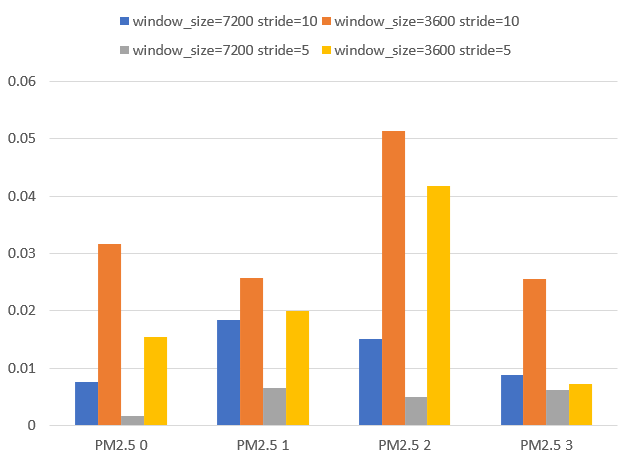}
    \caption{Window size 7200 vs 3600.}
    \label{fig:window_size3600_train_mse}
\end{figure}

Being inspired by the SOTA ideas of predicting the target sequence with a short sequence by using an auto-regression model such as Autoformer \cite{wu2021autoformer}, we approach to decrease the model's input size, i.e., the data's window size. We set the window size to 3600 (which means one hour on the timeline), and train our model again. Figure \ref{fig:window_size3600_train_mse} shows our results. Empirically, the model gains well performance as long as it reduces the training loss under 0.01. Hence, except for the result on \text{$PM_{2.5}$(3)} when the window size is set to 3600, the model still needs optimization if we want a shorter window size. However, it is worth trying as the number of model parameters also decreases obviously as the input size is reduced. A light model saves computational costs and boosts inference.

\section{Conclusion}

The WaveNet model designed for audio data processing is generalizable and suitable for fitting problem. Our work successfully put it into usage for IAQI level fitting and prediction. It shows that our GreenEyes model based on WaveNet has strong data fitting capability for extreme long data sequences. When given a smaller stride, fed with more data, the model can learn better. It is also found that, when trained with more channels of sensor data, the model can perform well. This can be regard as sensor data augmentation. Our innovative method that human manually label the IAQI level is useful. It creates an appropriated target label function that the model can learn and solve the threshold fluctuation problem.




It is also promising that our GreenEyes AIoT deployment design can be put into practice. Actually we've developed an iOS app to retrieve the air quality data. Mobile framework such as Tensorflow Lite \cite{louis2019towards} has been developed. A mobile phone is hopefully to be installed with our GreenEyes model and monitor the IAQI data in realtime and predict the air trend.

Due to a lack of air quality data, we only did the data fitting task. We will perform the data predicting task in the future if enough data is gathered.

\section{Related Works}
\subsection{Statistical \& Machine Learning Approaches}

Except for ARIMA, ETS models mentioned in our last chapter, traditional methods such as Kalman filter \cite{gomez1994estimation} are also very simple and practical for time series and forecasting problems. Random forests \cite{rouet2017machine}, XGBoost, and SVM \cite{sapankevych2009time} etc are useful machine learning methods too. About method choosing, the most suitable method is highly interrelated with the data's properties and the application scenario. 

In common, the essential of both traditional approaches and ML-based approaches is mining data and extracting features. Different from other feature engineering tasks, sliding windows are widely used for processing the data. Metrics such as the minimum, the maximum, the mean, and the variance of the data in the window are common features.

\subsection{Deep Learning Approaches}

LSTM-based deep learning methods have been developed recently to extract temporal patterns. Lai et al. proposed LSTNet \cite{lai2018modeling} that encodes short-term local information into low dimensional vectors using 1D convolutional neural networks and decodes the vectors through an RNN. Shih et al. proposed TPA-LSTM \cite{shih2019temporal}  which processes the inputs by an RNN and employs a convolutional neural network to calculate the attention score across multiple steps.

The architecture of CNN is designed for 2D data like images. Meanwhile, recently a special variant of CNN called temporal convolutional networks (TCNs) \cite{lea2016temporal} has been proposed that makes CNN capable for time series processing. Yan et al. \cite{yan2020temporal} released their research work about using TCN for weather forecasting in 2020 and showed that TCN is better than the LSTM network in this application.

WaveNet related methods, including our GreenEyes model, tackle with a single sequence of time series data and show good fitting and forecasting performance concerning the prediction accuracy and data throughput capacity. Meanwhile, same with the same recent time as this thesis was being developed, new methods and approaches regarding time series forecasting have also been proposed. In recent years, graph neural networks (GNNs) have shown high capability in handling relational dependencies. Wu et al. \cite{wu2020connecting} proposed a general graph neural network framework designed specifically for multivariate time series data. Their method is useful for extracts relations among variables belonging to multi sequences.

As Transformer \cite{vaswani2017attention} becomes great popular these years, another model based on Transforms has also been brought out. Lim et al. \cite{lim2021temporal} from Google introduced the Temporal Fusion Transformer (TFT) as a novel attention-based architecture which combines high-performance multi-horizon forecasting with interpretable insights into temporal dynamics. They created gate-based networks, GRN and GLU, as new approaches for better feature selection modules.

\begin{acks}
The authors would like to thank many friends for constructive discussions and feedbacks. Special thanks to \href{https://www.math.hkust.edu.hk/people/faculty/profile/yuany/}{Prof. Yuan Yao} who voluntarily provides GPU machine.
\end{acks}

\bibliographystyle{ACM-Reference-Format}
\bibliography{citations}

\end{document}